%% file: main.tex
\documentclass[final]{nesy2026} % Anonymized submission for Openreview
\usepackage{longtable}% for long tables
\usepackage{multirow}
\usepackage{graphicx}
\usepackage{bm}
\usepackage{pifont}
\usepackage{pgfplots}
\pgfplotsset{compat=1.18}
\usepackage[table]{xcolor}
\usepackage{tikz}
\usepackage{tcolorbox}
\usepackage{fvextra}  % adds wrapping to verbatim
\usetikzlibrary{calc,arrows.meta,positioning, shapes.geometric}
\usepackage{svg}
\usepackage{verbatim}
\usepackage{booktabs}
\usepackage{subcaption}
\usepackage[]{mdframed}
\usepackage[load-configurations=version-1]{siunitx} % newer version
\theorembodyfont{\upshape}
\theoremheaderfont{\scshape}
\theorempostheader{:}
\theoremsep{\newline}

\title[NeSyQuaKE]{A Neurosymbolic Approach 
for Explainable\\Early Diagnosis of Alzheimer's Disease}

  \author{\Name{Ranveer Singh\nametag{\thanks{R Singh, P Tenali and S Mathur - equal contribution}}} \Email{ranveer.singh@utdallas.edu}\\
   \Name{Pranuthi Tenali} \Email{pranuthi.tenali@utdallas.edu}\\
   \addr The University of Texas at Dallas, Richardson, TX, USA
   \AND 
   \Name{Saurabh Mathur} \Email{saurabh.mathur@tu-darmstadt.de}\\
   \addr  Technische Universität Darmstadt, Darmstadt, Germany
   \AND 
   \Name{Ameet Soni} 
   \Email{asoni1@swarthmore.edu}\\
   \addr Swarthmore College, Swarthmore, PA, USA
   \AND
   \Name{Vaishali Phatak} 
   \Email{vaishali.phatak@unmc.edu}\\
   \Name{Karla Lynch} 
   \Email{kalynch@unmc.edu}\\
   \Name{Daniel Murman} 
   \Email{dlmurman@unmc.edu}\\
   \Name{Matthew Rizzo} 
   \Email{matthew.rizzo@unmc.edu}\\
   \addr University of Nebraska Medical Center, Omaha, NE, USA
   \AND
   \Name{Sriraam Natarajan} 
   \Email{sriraam.natarajan@utdallas.edu}\\
   \addr The University of Texas at Dallas, Richardson, TX, USA}

\begin{document}

\maketitle

\begin{abstract}
Identifying reliable Alzheimer's disease (AD) markers typically requires manual, labor-intensive transcription and expert analysis, limiting its scale. We introduce an automated pipeline that extracts qualitative knowledge about potential AD progression indicators directly from audio recordings of verbal fluency tests. Our method uses pretrained foundation models to process raw audio and extract clinically relevant variables to construct a Bayesian Network (BN); this BN is used to reason about the AD progression markers and infer their qualitative relationships. Our system successfully recovers known clinical knowledge and identifies novel relationships between linguistic markers.
\end{abstract}

\section{Introduction}
Alzheimer's disease (AD) is a leading cause of mortality among elderly populations, currently ranking as the fifth-leading cause of death for Americans over the age of 65~\citep{adfactsandfigures2024}. With prevalence projected to double by 2060, there is an urgent need for scalable methods to detect early cognitive symptoms. These symptoms often manifest as subtle linguistic and semantic deficits~\citep{foster2013differential}, including reduced verbal fluency~\citep{marra2021semantic} and increased word-finding difficulty~\citep{salmon2011neuropsychological}. Clinical practitioners typically evaluate these markers through standardized neuropsychological assessments, such as verbal fluency tests~\citep{henry2004verbal,marczinski2006category}. However, the analysis of these tests remains a manual, labor-intensive bottleneck that requires expert transcription and qualitative analysis, severely limiting widespread longitudinal tracking.

The fundamental challenge in automating this discovery process lies in the \textbf{representational gap between sensory input and clinical knowledge}: cognitive impairment indicators are embedded within unstructured, continuous acoustic signals; clinical domain knowledge is structured around qualitative relationships between discrete variables. While neural architectures excel at perception and pattern recognition, they lack explicit mechanisms to ensure logically consistent, verifiable, and interpretable reasoning. Conversely,  symbolic systems cannot process raw, high-dimensional audio. Bridging this gap needs a framework capable of both high-fidelity symbol grounding and rigorous reasoning. 

To this effect, we introduce NeSyQuaKE (Neurosymbolic Qualitative Knowledge Extraction). Following a Neuro $|$ Symbolic (Type 3) architecture~\citep{kautz2022third}, we use pretrained foundation models for symbol grounding and Bayesian Networks for reasoning about pairwise relationships between cognitive impairment and candidate markers. NeSyQuaKE identifies the nature of the relationships between markers, represented as {\em Qualitative Influence Statements (QIs)}\citep{kuipers1994qualitative, karanam2021probabilistic,yang2013knowledge,odom2018human}. These QIs concisely capture trends such as monotonicity, allowing easy communication and validation of discovered knowledge against medical literature.

{\looseness -1 We make the following key contributions:
(1) We propose the NeSyQuaKE Framework, a novel neurosymbolic architecture that extracts qualitative knowledge from clinical audio using foundation models and Bayesian Networks; (2) We introduce a layer of deterministic symbolic programs to map structured neural outputs to clinical features, ensuring that the derivation of metrics such as speech rate and semantic switching is mathematically consistent and verifiable; (3) We evaluate our system on a real-world dataset of clinical verbal fluency tests, demonstrating that NeSyQuaKE successfully recovers established clinical knowledge and generates additional hypotheses about qualitative relationships between cognitive impairment and its markers.

}
% % This is pasted from UNMC document for reference
% Alzheimer’s disease (AD) is the fifth-leading cause of death in the US for people over 65.1 While the risk for AD increases with age, AD is not a normal aging process. Approximately 6.9 million Americans over the age of 65 suffer frompercentagepercentage Alzheimer’s disease and it is expected to grow to 13.8 million by 2060.1 Early cognitive symptoms of Alzheimer’s disease include rapid forgetting of newly learned material2 as well as language and semantic knowledge deficits such as reduced verbal fluency for items belonging to a semantic category,3 word finding difficulty,4 decreased low frequency word usage in semantic tasks,5 and longer pauses while speaking.6

% Retrieval of semantic knowledge from an individual’s knowledge store involves an intricate coordination of multiple brain systems. While an individual would have a unique knowledge base based on their educational and life experiences, the current neuropsychological practice for measuring semantic retrieval uses standardized cognitive tests, such as correctly naming a single visual or auditory stimulus in a prescribed time or producing a maximum number of examples from a specific semantic category (e.g., animals) within a prescribed amount of time.

% The primary goal of this project is to develop and validate an extraction of the language markers defining an individual’s semantic topology. This will assist in providing an early diagnostic marker as well as tracking disease progression.

% We make the following contributions: .
\vspace{-1em}
\section{Background}
\begin{figure}[t!]
    \centering
    \includegraphics[width=0.85\linewidth]{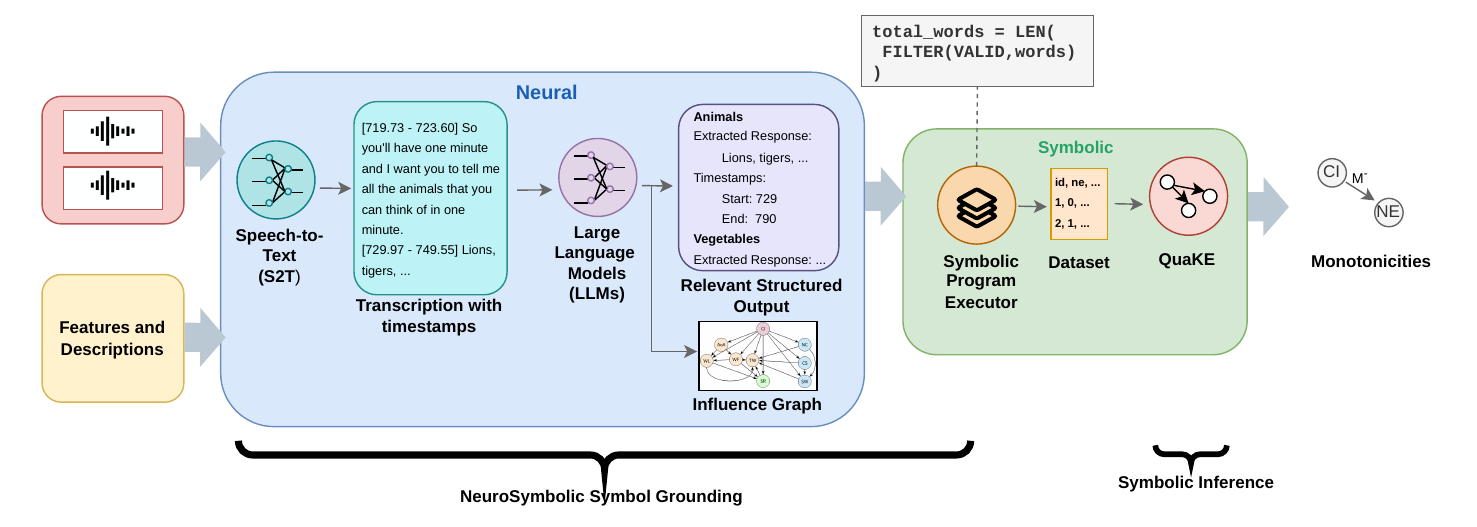}
    \caption{\footnotesize \textbf{End-to-end pipeline for automated discovery of qualitative AD markers.} Our framework extracts qualitative knowledge from raw audio data by combining foundation models with symbolic reasoning. It uses foundation models for robust transcription, information extraction, and generating an influence network. These data are combined with the influence network to instantiate a Bayesian Network (BN). The QuaKE algorithm uses this BN to infer underlying monotonicities between linguistic markers and disease progression.
    % This approach provides clinicians with interpretable qualitative insights without requiring manual feature engineering.
    }
    \label{fig:flowchart}
\end{figure}
%We now present the necessary background.
\paragraph{Neurosymbolic AI for Knowledge Discovery}\label{b:nesy}
Neurosymbolic AI~\citep{garcez2023neurosymbolic,marra2024statistical} integrates effective representation learning of neural networks with the structured reasoning of symbolic logic. 
% While pure neural architectures excel at pattern recognition, they lack explicit mechanisms ensuring that their internal representations are logically consistent and verifiable against existing domain knowledge. In contrast, a neurosymbolic approach enables rigorous inference by replacing heuristic black-box associations with formal reasoning engines, making it particularly well-suited for knowledge discovery.
The integration of these two paradigms is often categorized based on the interaction between their neural and symbolic components.  Our setting naturally aligns with the $\text{Neuro} \mid \text{Symbolic}$ (Type 3) system from Kautz's taxonomy~\citep{kautz2022third}, because our task of clinical knowledge discovery from audio clips requires both pattern recognition and strict, verifiable symbolic reasoning. % For instance, Kautz's taxonomy~\citep{kautz2022third} considers six types of systems, from Symbolic Neuro symbolic (Type 1), where neural networks are used within a symbolic framework, to Neuro [Symbolic] (Type 6), where symbolic logic is embedded within a neural architecture. The choice of architecture depends on each domain's requirements.
% Clinical knowledge discovery requires interpretable outputs, based on reliable reasoning and grounded in empirical data. However, the verbal fluency data considered in this work consists of unstructured audio data, which cannot be processed by a symbolic system alone. Neural network architectures are well-suited to process the raw audio and extract structured representations, i.e., symbols. Consequently, our setting naturally aligns with the Type 3 neuro-symbolic system from the taxonomy.
% We follow the $\text{Neuro} \mid \text{Symbolic}$ (Type 3) approach. Here, the neural and symbolic components are kept distinct: a neural perception module serves as a front-end that passes its output to a separate symbolic reasoner.
% Symbol grounding problem: the process of reliably mapping raw, high-dimensional sensory data into a structured symbolic space where formal reasoning can occur.
Instantiating such a system requires three critical components: \\
(1) \textbf{A Symbol Grounder:} A mechanism to map raw, high-dimensional observational data into discrete symbols. We address this symbol grounding problem by using Pretrained Foundation Models to extract clinical variables from noisy audio.\\
(2) \textbf{A Symbolic Language:} A formal representation to encode domain entities and their relationships. We use Bayesian Networks (BNs) as our formal language.\\
(3) \textbf{An Inference Procedure:} A mathematically sound method to derive new knowledge from the grounded symbols. We employ the QuaKE algorithm to perform this reasoning.

While symbol grounding has been addressed through learnable neural modules~\citep{hsu2023s}, the paucity of clinical data and the high cost of annotation make this less suited for medical domains. A promising alternative is to use pretrained foundation models~\citep{wust2025synthesizing}. These models are effective at extracting information from unstructured data such as images. To this effect, we use pretrained speech and language foundation models for intermediate symbol grounding and combine these with symbolic programs encoding domain knowledge to reliably map verbal fluency test recordings to variables of interest. However, instead of learning predictive models for cognitive impairment, we aim to infer explainable qualitative rules about the variables of interest and the nature of their relationship with cognitive impairment.

\paragraph{Foundation Models} 
% as Information Extractors}
\label{b:fm}
Foundation Models can provide a robust solution to the symbol grounding problem by acting as high-fidelity neural predicates or feature extractors.
Unlike traditional machine learning models trained on task-specific datasets, foundation models are pretrained on large corpora, enabling them to capture complex semantic and acoustic patterns required to extract clinically relevant information. We consider two types of these models: Speech-to-Text (S2T) and Large Language Models (LLMs).

Speech-to-text (S2T) models are a subclass of speech foundation models typically built on the transformer~\citep{vaswani2017attention} architecture that learn rich acoustic and semantic representations from large-scale audio data, specialized for tasks such as automatic speech recognition (ASR)~\citep{asr_survey}. For low-data settings such as clinical ASR where training or fine-tuning these models is not possible, pre-trained transformer-based models such as Whisper~\citep{radford2023robust} have been successful~\citep{hwang2025evaluating}.

Large language models (LLMs) are a class of foundation models trained on large corpora of text and used for natural language processing tasks such as question answering, summarization, and information extraction~\citep{10.1145/3744746, xu2024large}. Particularly
in clinical settings, domain-specific pre-trained LLMs such as MedGemma have been successful on information extraction tasks~\citep{zhou2025high}. 
However, using LLM outputs for symbolic grounding requires the outputs to conform to a predefined schema or format. This is often achieved through techniques like constrained decoding~\citep{wang2025slot}.

\paragraph{Bayesian Networks}\label{b:bn}
Bayesian Networks (BNs)~\citep{pearl1988probabilistic} are a powerful framework for representing and reasoning under uncertainty. They are a type of Probabilistic Graphical Models~\citep{koller2009probabilistic} that use Directed Acyclic Graphs (DAGs) to factorize the joint distribution over a set of variables $\bm{X}$ as a product of local conditionals. Specifically, a BN over $\bm{X}$ is defined as $\langle G, \bm{P} \rangle,$ where $G=(\bm{X}, E)$ is a DAG, consisting of nodes corresponding to each random variable and directed edges $E$ between them indicating direct influence, and each $P_X \in \bm{P}$ is the local conditional over $X$ given its parents $\text{pa}_X$. The joint probability of a data point $\bm{x}$ is factorized as $P(\bm{x}) = \prod_X P_X(x\mid \text{pa}_X(\bm{x})).$
Learning these joint distributions involves estimating the parameters of each local conditional. Maximum Likelihood Estimation (MLE)~\citep{Mitchell1997} estimates the parameters that maximize the likelihood of the data $\mathcal{D}, \hat{\theta}_{MLE} = \text{arg max}_\theta P(\mathcal{D|\theta})$. However, it fails for small data regimes where unobserved data configurations are assigned zero probability, thereby making such data points impossible under the model. A Bayesian Estimator~\citep{koller2009probabilistic} mitigates this issue by placing a prior such as BDeu~\citep{heckerman1995learning} over the parameters, $P(\theta)$, and computing a full posterior $P(\theta|\mathcal{D}) \propto P(\mathcal{D}|\theta)P(\theta)$. 
% A convenient choice for such priors are Dirichlet priors, such as the Bayesian Dirichlet equivalent uniform (BDeu)~\citep{heckerman1995learning} prior, which are conjugate priors to the categorical distribution, whose parameters $\alpha_{x|pa_X} = \frac{N'}{q_X r_X}$ can be interpreted as pseudo-counts where $N', q_X, r_X$ are the user-specified equivalent sample size, number of parent configurations and number of states of $X$. Moreover, setting $\alpha = 1$ yields a Laplace prior, while noisier data requires stronger priors, i.e., larger pseudo-counts, for stronger regularization.

%These models can be used to reason about the presence and the nature of relationships between subsets of variables via inference algorithms like Variable Elimination. For instance, the QuaKE algorithm uses BNs to infer Qualitative Influence Statements.

% Type 3 Neurosymbolic system per Kautz’s taxonomy
% \begin{quote}
%     ``[T]he risk of coexisting DU [duodenal ulcer] or GU [gastric ulcer] increases with an increase in grade of AG [atrophic gastritis] of the antrum but decreases with an increase in grade of AG of the gastric body''~\citep{sipponen1989chronic}
% \end{quote}

% \begin{quote}
%     ``[T]here was a monotonic increase in the relative risk of schizophrenia as paternal age progressed, leading to a 3-fold excess of the disease in offspring for whom paternal age was 50 years or more, relative to those of paternal age less than 25 years.''~\citep{malaspina2001advancing}
% \end{quote}
\paragraph{Domain Knowledge as Qualitative Influence Statements}\label{b:qis}
Qualitative Influence Statements describe the direction of and nature of interaction between variables~\citep{wellman1990fundamental}.
They are a common formalism in scientific knowledge, particularly medicine, enabling clinicians to communicate trends without specifying precise values~\citep{sipponen1989chronic,malaspina2001advancing}. We focus on a specific type of qualitative influence statement called a Monotonic Influence Statement, generalizing statements of the form ``$X$ increases as $Y$ increases.'' Such monotonic influence statements can be formalized using probabilistic logic to describe patterns in the conditional probability distribution~\citep{altendorf2005learning}. 

In a bivariate system, the statement that $Y$ positively monotonically influences $X$ implies that higher values of $Y$ lead to stochastically higher values of $X$. Formally, this corresponds to a first-order stochastic dominance or a leftward shift in the cumulative distribution:
$P(X \leq x \mid Y = a) \leq P(X \leq x \mid Y = b)\ \forall a > b, a, b \in \text{Domain}(Y)^2.$
% I don't think this line is relevant here: An additional `slack' parameter $\epsilon$ allows for a small tolerance to accommodate minor statistical fluctuations, preventing spurious violations of the constraint.
On the other hand, in multivariate systems, a qualitative influence can be defined either locally (holding all other parents of X constant~\citep{altendorf2005learning}) or marginally (marginalizing over all other factors~\citep{karanam2021probabilistic,karanam2025unified}). In this work, we take the latter approach. We define the influence of patient cognitive impairment $Y$ on a linguistic marker $X$ based on the posterior distribution $P(X\mid Y)$, obtained by marginalizing out all other markers $\bm{X} \setminus \{X\}$, allowing us to identify the global trends learned by the model from clinical data.
% Recapitulate the motivation and background.
% To do that, we first need the motivation and background finished
% in the form of QIs from the audio recordings of verbal fluency tests. 
\vspace{-0.5em}
\section{Neurosymbolic Qualitative Knowledge Extraction}

We consider the task of extracting qualitative influence statements describing the relationship between cognitive impairment and verbal fluency indicators, using a database of audio recordings. We formalize this as the following task:\vspace{-0.5em}
%\resizebox{\linewidth}{!}{
\begin{mdframed}

    \textbf{Given:} A dataset $\mathcal{D} = \{(a^{(i)}, y^{(i)})\}_{i=1}^N,$ where each $a^{(i)}$ is a raw audio recording and $y^{(i)} \in \{0,1\}$ is a diagnostic label indicating the presence of cognitive impairment;
    % collection of clinical audio recordings of fluency tests $A = \{a_1,\dots,a_N\}$, labeled with presence of cognitive impairment $Y, $ and 
    a set of clinical linguistic features $\noindent\bm{X} = \{X_1,\dots,X_n\}$\\
    \textbf{To Do:} Identify qualitative influences between the presence of cognitive impairment, $Y,$ and linguistic features $\bm{X}$

\end{mdframed}

    % Dataset containing audio recordings of each subject's fluency test, $\mathcal{A},$ the corresponding diagnostic labels $,Y,$ indicating the presence of cognitive impairment, and a list of variables of interest $\bm{X}$ \\
    % Extract the qualitative influence statements between Alzheimer's and its different markers
The key challenge in this problem is the heterogeneity of the data representations. On one hand, the cognitive impairment indicators of interest are embedded in unstructured, continuous acoustic signals $a^{(i)}.$ On the other hand, clinical knowledge is expressed through qualitative relationships between discrete variables $\bm{X}$ and $Y.$
Extracting this knowledge requires a system to perceive the nuances of natural language and reason about the probabilistic dependencies within a formal symbolic framework.

To bridge this gap, we introduce Neurosymbolic Qualitative Knowledge Extraction (NeSyQuaKE). As outlined in Figure~\ref{fig:flowchart}, NeSyQuaKE instantiates a $\text{Neuro} | \text{Symbolic}$ architecture to decouple the task into three distinct stages: Information Extraction (IE), Bayesian Network Construction (BNC), and Qualitative Knowledge Extraction (QuaKE).

% The framework used to solve the task of extracting qualitative influence statements from raw audio files, Neurosymbolic Qualitative Knowledge Extraction (NeSyQuaKE),  is shown in Fig.~\ref{fig:flowchart}. The framework consists of the following stages: Information Extraction (IE), Graph Construction (GC), and Qualitative Knowledge Extraction (QuaKE), with the stages described in the subsequent subsections

\subsection{Information Extraction}
\begin{table}[!t]
    \centering
    \input{Tables/features}
\end{table}

The Information Extraction (IE) stage maps each audio recording $a^{(i)}$ to a tabular feature vector $\bm{x^{(i)}},$ transforming the original dataset $\mathcal{D}$ into a tabular dataset $\mathcal{D}' = \{\bm{(x^{(i)}}, y^{(i)})\}_{i=1}^N,$  making it suitable for Bayesian Network-based reasoning.
% takes the set of raw audio files ${A}$, and outputs the structured responses extracted from the audio files ${R}$, which can be further processed to extract a tabular dataset ${D}$. 
It does so using the \textbf{Speech to Text (S2T)} and the \textbf{Large Language Models (LLM)} modules sequentially.

The \textbf{S2T} module takes each raw audio file $a^{(i)}$ as input and outputs the corresponding transcriptions $t^{(i)}$, each consisting of a set of timestamped utterances. That is,
$t^{(i)} = \text{S2T}(a^{(i)})\ \forall i = 1,\dots,N\text{ where } t^{(i)} = \{(t^{(i)}_1, u^{(i)}_1), \dots,(t^{(i)}_{k_i}, u^{(i)}_{k_i})\}$
We implement \text{S2T} using a pre-trained transformer-based foundation model trained for speech recognition. Next, we use these transcriptions ${T} = \{t^{(i)}\}_{i=1}^N$ to create the tabular dataset by aligning them with the variables of interest, $\bm{X}$.

% are not structured in a way conducive to automatic parsing and processing to construct the tabular dataset $\mathcal{D}$ required for downstream tasks.
The \textbf{LLM} module constructs the tabular datasets by using 
%To construct the tabular dataset required for downstream tasks, the Large Language Model (LLM) module takes 
the transcripts $T$ as input along with a custom prompt ${P}$ describing the scope of information required, extraction rules, and constraints.  The LLM outputs a structured response ${R}$, 
%, which can be parsed and processed automatically. Specifically, the feature extractor is implemented by invoking a pre-trained decoder-based large-language model with syntactic constraints $S$, which outputs a structured response ${R}$ as specified by $S$. % $\mathbf{LLM}: {T, P}, S \rightarrow {R}$,  
% This can then be processed automatically to construct the final dataset ${D}$. 
which serves as a grounded intermediate representation, where the LLM has mapped the raw transcript $t^{(i)}$ into a structured sequence of words and temporal markers. However, these individual grounded symbols do not directly correspond to the variables of interest required for BN reasoning.

To complete the grounding, we define a set of \textbf{deterministic symbolic programs} to compute the feature vectors $\{\bm{x}^{(i)}\}_{i=1}^N$. This step ensures that every clinical variable is derived through a transparent and consistent calculation. For example, while a dictionary lookup function identifies which words are valid responses, a simple length function determines the total words (TW), i.e,  \texttt{TW = LEN(FILTER(valid\_words, words))}. Similarly, the speech rate is calculated by dividing the number of identified words by the total duration recorded in the temporal markers. This ensures that the resulting dataset is logically consistent while being grounded in the information extracted by the foundation models.
% \begin{table}[ht]
%     \centering
%     \begin{tabular}{l l p{8cm}}
%     \toprule
%     \textbf{Type} & \textbf{Feature} & \textbf{Pseudo-code} \\
%     \midrule
%     \rowcolor{orange!10} \textbf{Lexical} 
%         & Number of Exemplars  & \texttt{LEN(FILTER(VALID, words))} \\
%     \rowcolor{orange!10} & Word Frequency       & \texttt{MEAN(MAP(CORPUS\_FREQ, FILTER(VALID, words)))} \\
%     \rowcolor{orange!10} & Word Length          & \texttt{MEAN(MAP(LEN, FILTER(VALID, words)))} \\
%     \rowcolor{orange!10} & Age of Acquisition   & \texttt{MEAN(MAP(AOA, FILTER(VALID, words)))} \\
%     \midrule
%     \rowcolor{cyan!10} \textbf{Semantic} 
%         & Number of Clusters   & \texttt{LEN(DISTINCT(MAP(CLUSTER\_ID, words)))} \\
%     \rowcolor{cyan!10} & Average Cluster Size & \texttt{(LEN(words) - 1) / LEN(DISTINCT(MAP(LUSTER\_ID, words)))} \\
%     \rowcolor{cyan!10} & Number of Switches   & \texttt{LEN(FILTER(switched, ZIP(MAP(CLUSTER\_ID, words), MAP(CLUSTER\_ID, words[1:]))))} \\
%     \midrule
%     % \rowcolor{green!10} \textbf{Acoustic} 
%     %     & Number of Pauses     & \texttt{LEN(FILTER($\lambda$w: w.start - prev(w).end > threshold, words))} \\
%     \rowcolor{green!10} \textbf{Acoustic} & Speech Rate          & \texttt{LEN(words) / (words[-1].end - words[0].start)} \\
%     \bottomrule
% \end{tabular}
% \caption{Symbolic programs used to compute values for Lexical, Semantic, and Acoustic variables from LLM response}
% \label{tab:features_pseudocode}
% \end{table}

\subsection{Bayesian Network Construction}
The Bayesian Network Construction (BNC) stage is responsible for constructing the influence graph (a BN) over the variables of interest $\bm{X} = \{X_1, X_2, \dots, X_n\}$ as well as learning the joint probability distribution $P(\bm{X}) = P(X_1, X_2, \dots, X_n)$ over the graph $G = (\bm{X}, E).$ Since reliable structure learning is infeasible in domains with small and noisy data, we obtain the initial graph $G$ over the variables from an LLM by prompting it multiple times with the task and feature description to output a BN structure.
The resulting graphs are pooled to construct the final graph by adding edges in decreasing order of occurrence, skipping any edges that introduce cycles to maintain the acyclicity constraint. Next, the dataset generated in the IE stage is used to learn the joint distribution $P(\bm{X})$ for the BN. Due to the stochastic nature of the different components of the IE stage and the noise inherent in raw audio files used for dataset construction $\mathcal{D'}$, a Bayesian Estimator with a strong prior like BDeu is used to regularize the parameter estimates.

% Need either a better abbreviation or a new name for the features
\subsection{Qualitative Knowledge Extraction}

% The dataset constructed in the IE stage, as well as the influence graph over the features elicited from domain experts, are used as input to the Qualitative Knowledge Extraction (QuaKE) stage. The QuaKE module extracts the qualitative influence statements, mainly monotonicities and synergies, over these features.

% Read the QuaKE paper again and more or less summarize it and its main formulations

% This is taken mostly from section 2.1 of the QuaKE paper. It might make more sense to move some of the definitions for monotonicity and synergies into the background section.
The Qualitative Knowledge Extraction module is based on the QuaKE algorithm~\citep{karanam2021probabilistic}. It reasons over the joint distribution $P(\mathbf{X}, Y)$ of Alzheimer's diagnostic labels $Y$ and the discretized feature vector $\mathbf{X}$ to extract the qualitative knowledge as Monotonic Influence (MI) statements. 
For each descendant $X_d$ of $Y$, QuaKE checks for the presence of monotonic influence of $Y$ on its descendant $X_d$ as 
$$ C_d = \prod_{j}\prod_{j'}\prod_k \text{max}(P(X_d \leq k | Y = y^j) - P(X_d \leq k | Y = y^{j'})  + \epsilon, 0)$$ where $\epsilon$ is the monotonic slack that allows a small tolerance for violation of this constraint.
{\looseness -1
A positive MI ($Y \prec^{M^+} X_d$) requires that higher values of $Y$ make it less likely that $X_d$ takes smaller values, which implies that $\forall_{k, j' > j}$, $P(X_d < k|Y= y^j) \geq P(X_d < k|Y = y^{j'})$. Essentially, $C_d$ is non-zero if and only if the monotonicity constraint holds for all the values of $Y$, i.e., for all $(y^j, y^{j'})$ where $j'> j$ and for all the values of $k$. The same holds for negative monotonicities.
When the monotonicity constraint is satisfied, the degree of monotonic influence $\delta_d$ is then computed as 

}
\begin{equation}\label{eq:monotonic}
\delta_d = I_{C_d > 0} \sum_{j}\sum_{j' > j}\sum_{k} \frac{P(X_d \leq k | Y = y^j) - P(X_d \leq k | Y = y^{j'}) }{|Y|}
\end{equation}
where $\delta_d $ quantifies the strength of the monotonic influence. Intuitively, stronger monotonic influences result in larger cumulative differences between the conditional probabilities $P(X_d \leq k | Y = y^j)$ and $P(X_d \leq k | Y = y^{j'})$ across the label values and threshold $k$. So, larger absolute values of $\delta_d$ represent a stronger monotonic influence. In addition to the MIs of $Y$ on its descendants, this module also extracts the positive and negative monotonic influences of each feature $X_i$ on its descendants in the graph.

\section{Empirical Evaluation}
We now present our empirical evaluations\footnote{The code for all the experiments 
% including the NeSyQuaKE (w/wo) feedback and the baseline implementations,
is available in the  \href{https://github.com/s-ranveer/verbal_fluency_alzheimers}{supplementary}} and answer the following questions:
% \\\textbf{Q1)} Is the Foundation Model-based pipeline effective for symbol grounding for verbal fluency data? 
\\\textbf{Q1)} Is NeSyQuaKE effective at symbol grounding of verbal fluency data? 
% How accurate are the pretrained Foundation models at symbol grounding component at extracting the data?
\\\textbf{Q2)} Does the knowledge extracted from the NeSyQuaKE pipeline match expert knowledge?
\\\textbf{Q3)} Is NeSyQuaKE an effective source of generating hypotheses about AD markers?
\\\textbf{Q4)} Can feedback between symbolic and neural components improve symbol grounding?

\textbf{Dataset:} To answer these questions, we consider a dataset of 162 audio files of verbal fluency tests from the University of Nebraska Medical Center, focusing on patients classified as Healthy Controls (HC) or Mild Cognitive Impairment (MCI). 
Manually administering these verbal fluency tests and diagnosing the patients is both time-consuming and labor-intensive. %, as it requires the clinicians to carefully review the answers provided by each patient. Consequently, the dataset is smaller in size. 
% However, the data is very rich %the limited sample size is compensated for by the high quality of the dataset %
Each patient in the study underwent a comprehensive neuropsychological battery annually for three years, and their diagnoses for each year were reviewed by the domain experts: two neurologists and a neuropsychologist. Since we aim to identify knowledge for early diagnosis, we use data from the first year.

Our primary focus is on patients’ ability to generate words within one minute, either beginning with specific letters (e.g., F or L) or belonging to a given semantic category (e.g., animals or vegetables), and evaluating the known clinical AD markers~\citep{ kave2016word,venneri2008anatomical,troyer1998clustering,pistono2019happens}
% Add references for the different markers here
described in Table~\ref{tab:features_table_colored}.
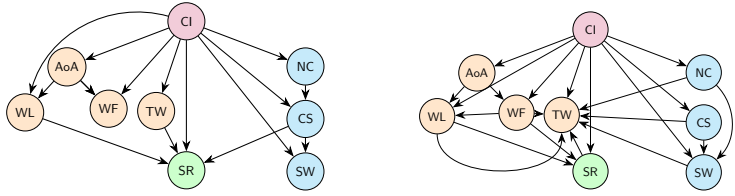
\begin{figure}[!t]
    \centering
    %\hfill 
    \begin{minipage}[t]{0.35\textwidth}
        \centering
        \resizebox{0.8\linewidth}{!}{\input{Figures/variable_bn.tex}}
        % \captionof{figure}{Expert Graph}
        % \label{fig:expert}
    \end{minipage}
    % \hfill
    \begin{minipage}[t]{0.35\textwidth}
        \centering
        \resizebox{0.8\linewidth}{!}{\input{Figures/llm_bn.tex}}
    \end{minipage}
    \hfill
    \caption{The \textbf{Influence Networks}, elicited from the expert (left), and LLM responses (right) representing direct influence relationships between \colorbox{purple!20}{CI}, \colorbox{orange!20}{Lexical}, \colorbox{cyan!20}{Semantic}, and  \colorbox{green!20}{Acoustic} variables.}
    \label{fig:bn_comparison}
\end{figure}
% For the Information Extraction (IE) stage of NeSyQuaKE, we use 

\textbf{Experimental Setup:} 
We instantiate NeSyQuaKE as follows: we use the WhisperX Speech to Text model and the MedGemma-27b medical LLM~\citep{medgemma} as the foundation models\footnote{Owing to the sensitive nature of medical data, only locally available LLMs were used for data processing}. We enforce syntactic constraints on LLM output using Pydantic\footnote{Without structural decoding, none of the outputs generated match the schema specification. Using structural decoding, on average 97.5\% of all outputs were schema compliant across 5 seeds.}, which are fed as input to symbolic programs to generate the final dataset\footnote{The symbolic programs, auxiliary data used (valid word lists, word groups, word frequencies, and age-of-acquisition data), structure learning, and transcription error analysis are provided in the appendix.} over the features described in Table~\ref{tab:features_table_colored}. Finally, to discretize the dataset, we use the mean feature values as a threshold\footnote{Threshold is computed as the mean over each variable across 5 different trials, as per our domain experts' recommendation, with the exact values provided in the supplementary}. For QuaKE, the LLM-generated BN
(Figure~\ref{fig:bn_comparison} (right)) is fed alongside the data to learn the parameters using a Bayesian Estimator with a BDeu prior and an equivalent sample size of 10 to account for noisy data and a monotonic slack of 0.
%Add the NeSyQuake setup afterwards

\textbf{Results:} 
Tables ~\ref{tab:pcee_comparison} and~\ref{tab:fluency_indicators} present the results for the questions introduced above.
\begin{figure}[!t]
    \centering
        \input{Figures/mi_bar_llm}
\end{figure}

\textbf{(Q1, Symbol Grounding)}
To evaluate the efficacy of our symbolic grounding module, we compare its output (i.e., variable values) to manually annotated data.
%the audio recordings with the words in each subject's response. We then obtained a ground truth dataset by applying the same symbol grounding executors and thresholds used in our framework. 
To quantify the resulting difference in pairwise relationships, we introduce the Pairwise Conditional Estimation Error (PCEE), which is computed as 
\begin{equation*}
PCEE(X|Y) = \frac{\sum_Y ({P}_{\text{NeSyQuaKE}}(X=1| Y =y) - {P}_{\text{manual}}(X=1| Y =y))^2}{|Y|}    
\end{equation*} where $Y$ denotes the cognitive impairment and $X$ represents a variable of interest. % PCEE computes the mean squared deviation between the conditional probability $P(X|Y)$ computed from NeSyQuaKE-grounded features and those from manual annotation. 
A higher PCEE (closer to 1) indicates a larger discrepancy, while a lower PCEE indicates close agreement between NeSyQuaKE groundings and manual annotations.
% Table \ref{tab:variables_combined} reports the thresholds and PCEE values for all the features across both the fluency tasks. Threshold is computed as the mean over each variable across 5 different trials, as per our domain expert's suggestion, and frequency is the number of data points where the value is greater than the threshold. 
We compare the PCEE of our symbol-grounding pipeline against a rule-based baseline that uses a set of hardcoded patterns and keywords for approximately parsing the transcript.
Table \ref{tab:pcee_comparison} shows the PCEE values for different methods. NeSyQuaKE obtains substantially lower PCEE than the domain-specific rule-based extractor, yielding an improvement of $94.3\%$ in the semantic fluency task and $77.2\%$ in the phonemic task, indicating higher reliability and match with manual annotations.
% This suggests that our framework accurately grounds verbal fluency data. While we observe relatively higher PCEE values for total words, word length, and speech rate in the semantic fluency task, and for age of acquisition and number of switches in the phonemic task, we attribute this discrepancy to missing words in the S2T outputs. Transcription performance degrades heavily in the presence of high background noise. This issue stems from the upstream transcription model and does not indicate poor symbol grounding of the NeSyQuaKE framework. 

% We extract the monotonic influences of 'Cognitive Impairment (CI)' on each of its descendants in the graph using \equationref{eq:monotonic}.
 % except the one between Cognitive Impairment (CI) and Word Length (WL) for the phonemic fluency task. Here, NeSyQuaKE 
% We don't have an edge between WL and TW. So, I am not sure what monotonicities hold here. This is what I am speculating based on Dr. Phatak's response~

\textbf{(Q2, Comparison with Expert)} 
% We evaluate the extent to which both the LLM-elicited graph and the monotonic influences extracted by NeSyQuaKE align with the expert knowledge.
% Figure \ref{fig:bn_comparison} shows the differences between the expert and the LLM-elicited graph. The LLM-elicited graph is comparatively denser than the expert graph. To quantify the differences between these two graphs, we compute the Structural Hamming Distance (SHD), precision, and recall. SHD measures the number of edge additions, deletions, or reversals needed to transform one graph into another. The SHD between the two graphs is $11$, indicating that the LLM-elicited graph substantially diverges from the expert graph. The precision is $0.57$, meaning that $57\%$ of the edges in the LLM-elicited graph are correct as they are present in the expert graph, while the recall being $0.8$ indicates that the LLM can recover $80\%$ of the edges from the expert graph.
Table \ref{tab:fluency_indicators} compares the monotonic influences extracted by NeSyQuaKE to those listed by our domain experts for the semantic and phonemic fluency tasks. For both tasks, NeSyQuaKE matched all influences initially elicited from the expert. In addition to the direction of influence, NeSyQuaKE also quantifies their degree of influence as shown in Figure \ref{fig:MI_degrees}. Additionally, we compare the LLM-generated Bayesian Network structure used by NeSyQuaKE to an expert-defined network. We find that the LLM-generated network recovers $80\%$ of the edges from the expert network. However, it contains several extra edges. $43\%$ of its edges were not present in the expert network. In contrast, the data-driven Peter-Clark (PC~\citep{pc}) algorithm yields a sparser network, matching only 1 out of 15 expert network edges. Overall, the LLM-generated Bayesian Network structure is conservative, encoding fewer independencies, which allows its data-driven parameters to recover patterns accurately.
% Specifically, it identified a positive monotonic influence of cognitive impairment on word length, which contradicted the experts. One possible explanation for this discrepancy is that the data were obtained from a fluency task rather than a spontaneous conversation. In these tasks, the patients are instructed to generate as many words as possible within a one-minute limit. To maximize their total words, patients may strategically use smaller words, increasing the total number of words at the cost of word length (WL). Such strategies are not possible in semantic fluency tasks where responses are constrained to fixed categories like vegetables and animals. Since NeSyQuaKE successfully recovered every other expert-provided monotonicity, we can answer \textbf{Q2} affirmatively.

\begin{table}[!t]
\centering
    \input{Tables/PCEE}
     
\end{table}

\textbf{(Q3, Additional Hypotheses)}
% Of the monotonic influences listed in Table \ref{tab:fluency_indicators}, 
Our framework discovers 9 additional influences beyond those initially elicited from the domain expert. Of these, 6 were in the phonemic task, and 3 were in the semantic task. In the semantic task, it infers the negative monotonic influence of cognitive impairment on cluster size, speech rate, and word length. These were validated by the domain expert and confirmed to be correct, since Healthy Control (HC) participants typically exhibit higher word length, larger cluster sizes, and higher speech rates than individuals with MCI in semantic fluency. For the phonemic task, our framework infers a positive monotonic influence of cognitive impairment on word length and a negative monotonic influence of cognitive impairment on total words, number of clusters, cluster size, number of switches, and speech rate. The experts concurred with the negative monotonic influences; while these patterns are known to exist in the phonemic task, the effect is typically less pronounced than in the semantic task~\citep{beatty1997influences,laws2010normal}. Nevertheless, NeSyQuaKE could detect these subtle patterns.
% Although semantic fluency is generally more consistent in discriminating HC from MCI, it is still plausible that healthy participants produce a higher number of clusters, larger average cluster sizes, more switches, and higher total word counts than cognitively impaired participants in phonemic tasks. These patterns and influences are comparatively more subtle than those observed in semantic fluency. 
% Nevertheless, NeSyQuaKE could infer such subtle monotonic relationships highlights the efficacy of our framework.  
{\looseness=-1 On the other hand, the experts were less certain about the positive influence of cognitive impairment on word length. We speculate that this could be a result of the setup of the phonemic task. In it, the participants were instructed to generate as many words as possible within a one-minute interval. To achieve this, participants may have strategically used smaller words to maximize the total word count at the cost of average word length. However, such strategies are not possible in semantic fluency tasks where responses are constrained to fixed categories like vegetables and animals.

}
\begin{table}[!t]
\centering

\input{Tables/monotonic_influences}
\end{table}

\textbf{(Q4, Incorporating Feedback)} 
We consider incorporating feedback from the symbolic component to the neural component. The symbolic component tests words for relevance to the specific fluency task and rejects irrelevant words, such as ones not beginning with the correct letter. However, these may represent mistranscriptions. We feed these rejected words to the LLM and prompt it for phonetically similar replacements relevant to the task; we keep the highest-confidence replacements to correct the mistranscriptions. Table~\ref{tab:pcee_comparison} shows that incorporating additional feedback resulted in an improvement over NeSyQuaKE by $13.7\%$ and $14.5\%$ in the semantic and phonemic tasks, respectively, but with the same set of monotonicities discovered as before. While the LLM correctly guesses the correct versions of mistranscribed words, this correction is not yet grounded in the audio clip, but is purely based on the transcript context. Therefore, while there may be a performance improvement using a feedback loop, further study is necessary to validate it.

\section{Conclusion and Future Work}
We introduced the NeSyQuaKE framework, a neurosymbolic framework designed to extract qualitative knowledge from clinical audio recordings of verbal fluency tests. It does so by employing a Neuro$|$Symbolic (Type 3) architecture, bridging the representational gap between unstructured acoustic signals and structured qualitative knowledge in the form of monotonic influence statements. Our empirical evaluation on a real-world dataset shows that NeSyQuaKE is effective at recovering established clinical domain knowledge. Moreover, it also inferred less prominent trends, like the negative monotonic influence of cognitive impairment on the number and size of clusters in the phonemic task.

There are several directions for future work. First, our choice of foundation models was limited by resource constraints and the sensitive nature of clinical data. Exploring larger foundation models and developing models specialized for processing pathological speech~\citep{cohn2026ASR} is important future work. Additionally, future work should explore Speech Language Models (SLMs) like SALMONN \citep{tangsalmonn} to enable tighter integration between speech processing and information extraction.
Second, future work should consider ways to scale the analysis to a larger and more complex set of variables of interest, such as the number of pauses, repetitions, and the rate of change in speech rate during the course of the response, and employing tractable probabilistic models.
Finally, while NeSyQuaKE identifies the nature of relationships for a given set of variables of interest, it can be extended to discover new AD markers, such as by using an LLM to \textit{generate} additional candidates and inferring the extent of cognitive impairment's influence on them. Together with these directions, NeSyQuaKE offers a scalable, explainable approach to tracking Alzheimer's disease progression and supporting early diagnostic efforts.

% is our first step towards our goal of building a fully automated pipeline for generating and testing hypotheses in a domain in the form of qualitative influence statements. Currently, NeSyQuake only considers the features provided by the domain experts. However, we envision a fully closed-loop where NeSyQuaKE is augmented to include a generator module (like an LLM), which can come up with features of interest by itself when provided the domain of interest, and augment the feature set itself based on the results by dropping irrelevant features, and selecting new features based on the best performing ones to help experts find new features to augment their research.
\acks{The authors gratefully acknowledge the support by AFOSR award FA9550-23-1-0239 and the Cluster of Excellence “Reasonable AI” funded by the German Research Foundation (DFG) under Germany’s Excellence
Strategy (EXC-3057).}
\clearpage
\bibliography{bibliography}

% \appendix
\newpage\noindent
\appendix
% \setcounter{page}{1}
% \begin{center}
%     \LARGE \textbf{Supplementary: NeSyQuaKE}
% \end{center}
% The appendix is organized as follows: In section A, we define the symbolic programs. Then, in section B, we provide the prompts given to the LLM for generating the structured output, as well as for obtaining the initial influence graph. In section C, we provide additional implementation details.
\section{Symbolic Programs}
\subsection{Pseudocodes}
Table \ref{tab:features_pseudocode} shows the list of features and the pseudocode of the deterministic symbolic programs that are used to extract these feature values. These symbolic programs take the structured JSON obtained from the LLM as input\footnote{The pseudocode table is for reference, with the actual implementation provided in the \texttt{construct\_features.py} file provided in the linked repository: \url{https://github.com/s-ranveer/verbal_fluency_alzheimers/blob/main/construct_features.py}}
\begin{table}[h]
    \centering
    \resizebox{.95\linewidth}{!}{\begin{tabular}{l l p{8cm}}
    \toprule
    \textbf{Type} & \textbf{Feature} & \textbf{Pseudo-code} \\
    \midrule
    \rowcolor{orange!10} \textbf{Lexical} 
        & Number of Exemplars  & \texttt{LEN(FILTER(VALID, words))} \\
    \rowcolor{orange!10} & Word Frequency       & \texttt{MEAN(MAP(CORPUS\_FREQ, FILTER(VALID, words)))} \\
    \rowcolor{orange!10} & Word Length          & \texttt{MEAN(MAP(LEN, FILTER(VALID, words)))} \\
    \rowcolor{orange!10} & Age of Acquisition   & \texttt{MEAN(MAP(AOA, FILTER(VALID, words)))} \\
    \midrule
    \rowcolor{cyan!10} \textbf{Semantic} 
        & Number of Clusters   & \texttt{LEN(DISTINCT(MAP(CLUSTER\_ID, words)))} \\
    \rowcolor{cyan!10} & Average Cluster Size & \texttt{(LEN(words) - 1) / LEN(DISTINCT(MAP(CLUSTER\_ID, words)))} \\
    \rowcolor{cyan!10} & Number of Switches   & \texttt{LEN(FILTER(switched, ZIP(MAP(CLUSTER\_ID, words), MAP(CLUSTER\_ID, words[1:]))))} \\
    \midrule
     \rowcolor{green!10} \textbf{Acoustic} & Speech Rate          & \texttt{LEN(words) / (words[-1].end - words[0].start)} \\
    \bottomrule
\end{tabular}}
\caption{Symbolic programs used to compute values for Lexical, Semantic, and Acoustic variables from LLM response}
\label{tab:features_pseudocode}
\end{table}
\newpage\noindent
\subsection{Data Sources}
The following provides the list of data sources that were considered for the task
\subsubsection{Word Frequency}
The Python library \texttt{wordfreq}\footnote{Refer to the \href{https://github.com/rspeer/wordfreq/?tab=readme-ov-file}{package repository}  for additional details} was used to compute the word frequency for the different words.
\subsubsection{Age Of Acquisition}
The Age of Acquisition is a 2017 corpus of over forty thousand different words~\citep{brysbaert2017test}. Any words considered when constructing the Age of Acquisition (AoA) feature; any word not in the list (no word or its lemma didn't match) were not considered during computation.
\subsection{Valid Groups}
The same animal and vegetable groups as in ~\citep{CLARK2014202} were used to group the animal and vegetable responses.  The animal list was adopted without modification, while the grocery list was filtered to include only vegetables by removing fruits, meats, and processed foods to get the final list of vegetables for semantic fluency word groups.

For the Phonemic Fluency, two words were said to belong to the same group if they started with the same two letters, were homonyms, rhymed, or differed by only a vowel sound.

The valid words are defined separately for the semantic and phonemic tasks. 
\begin{enumerate}
    \item \textbf{Semantic:} The words should belong to the group list of vegetables or animals to be considered valid
    \item \textbf{Phonemic:} The word must have a word frequency of greater than 0, not be a proper noun (name, person, or number), and should begin with the letter in question when asked to output a word list.
\end{enumerate}
Only valid words are considered for the metrics, except for Speech rate, where we need to consider all the words spoken.
\newpage\noindent
\section{Prompts}
The different prompts used with the LLM are provided below. The LLM used was Medgemma-27b with a total context window of 40,000 tokens 
% \subsection{Prompt for Information Extraction}

\begin{minipage}{\textwidth}
\begin{tcolorbox}[width=\textwidth]
\footnotesize
\begin{Verbatim}
## Alzheimer's Fluency Tests Processing
You are processing transcripts from a cognitive assessment for Alzheimer's disease. 
The transcript contains the timestamped utterances (questions and answers) from both the examiner and the participant.
We are interested in extracting the responses to the following standard verbal fluency prompts which would go something like:
1. "Let's begin. Tell me all the words you can, as quickly as you can, that begin with the letter 'F'. Ready? Begin." 
2. "Now I want you to do the same for another letter. The next letter is 'L.' Ready? Begin."  
3. "Now I want you to name things that belong to another category: Animals. You will have one minute. I want you to tell me all the animals you can think of in one minute. Ready? Begin."
4. "Now I want you to name things that belong to another category: Vegetables. You will have one minute. I want you to tell me all the vegetables you can think of in one minute. Ready? Begin."

The responses to these standard prompts should be around 60 seconds from the start of the response.

Your task is to extract the following from the transcripts:
1. The full response by the participant for the prompt ignoring the other speaker.
2. The extracted answer from the full response where one omits the filler words and incorrect answers. Repetitions are allowed.
3. The starting and ending timestamp of the participant response.
4. Individual pauses during the participant's response where the pause is at least 1 second, calculated as the time gap between the end timestamp of one utterance and the start timestamp of the next utterance.

**Rules**
1. Do not infer or reconstruct missing speech.
2. Return an empty string for "full_response" and empty arrays for other fields if no response exists for that task.
3. Output only the final JSON with no additional commentary or explanation.
4. Use R1, R2, R3, and R4 to represent the 4 task prompts respectively.
5. Timestamps should be preserved in their original format from the transcript.

**Edge Cases**
- If a participant self-corrects (e.g., "cat... no, dog"), include both in full_response but only the final word in extracted_answer if valid.
- If the examiner interrupts, note the interruption point as the end timestamp.
\end{Verbatim}
\end{tcolorbox}
\captionof{figure}{Prompt for Information Extraction (Part 1).}
\label{fig:prompt_ie_1}
\end{minipage}

\begin{minipage}{\textwidth}
\begin{tcolorbox}[width=\textwidth]
\footnotesize
\begin{Verbatim}
**Output JSON Schema**
```json
{
  "responses": {
    "R1": {
      "full_response": "string",
      "response_timestamps": {
        "start": "string",
        "end": "string"
      },
      "extracted_answer": ["word1", "word2"],
      "pauses": [
        {
          "start": "string",
          "end": "string"
        },
        {
          "start": "string",
          "end": "string"
        }
      ]
    },
    "R2": { },
    "R3": { },
    "R4": { }
  }
}
```

**Notes**
1. When we ask for the full response, we mean the entire answer provided by the participant including any errors. 
2. A pause occurs when there is more than a second difference between the end timestamp of one line and the start timestamp of the next line.
3. The extracted answer for R1 and R2 should not include names of people or places or numbers. 
4. Make a proper judgment of when the response begins and ends. Many times, there would be a sentence including phrases like "Stop here" to indicate the end of the response.
\end{Verbatim}
\end{tcolorbox}
\captionof{figure}{Prompt for Information Extraction (Part 2).}
\label{fig:prommpt_ie_2}
\end{minipage}

% \subsection{Prompt for Eliciting Corrections using Feedback}
\begin{minipage}{\textwidth}
\begin{tcolorbox}[width=\textwidth]
\footnotesize
\begin{Verbatim}
# Alzheimer's Verbal Fluency Test Corrections

You are a clinical expert reviewing verbal fluency responses extracted from speech transcripts.

Patients are elderly and may have mild cognitive impairment, dementia, speech disfluencies, articulation difficulties, or hearing impairments. Transcription and word extraction systems may also introduce recognition errors.

## Task

For each rejected word, determine whether it is a transcription or extraction error and identify the most likely intended word.

Rejected words failed symbolic grounding — they are either not valid words or do not belong to the expected response category.

Your goal is solely to identify likely transcription or extraction errors. Do not attempt to improve the patient's score or force words into the expected category.

## Response Categories

The input contains responses across four verbal fluency tasks:

- **R1** — Letter fluency: words beginning with the letter F
- **R2** — Letter fluency: words beginning with the letter L
- **R3** — Category fluency: Animals
- **R4** — Category fluency: Vegetables

## Correction Process

Apply the following steps in order for each rejected word.

### Step 1: Generate Phonetic Candidates

Identify all real words that are phonetically similar to the rejected word. Prefer candidates requiring minimal sound changes — ideally a single phoneme substitution, deletion, or insertion.

Be especially skeptical of candidates that change the initial phoneme — these represent larger phonetic departures and require stronger justification.

Never consider a candidate that is identical in pronunciation to the rejected word. No transcription error can exist between homophones.

### Step 2: Filter by Task Category

Eliminate any candidate that does not satisfy the task category constraint:

- For R1: candidate must begin with F
- For R2: candidate must begin with L
- For R3: candidate must be an animal
- For R4: candidate must be a vegetable

If no candidates remain after filtering, return no correction.

Note: category is used here only to eliminate impossible candidates, not to generate or prefer corrections.
\end{Verbatim}
\end{tcolorbox}
\captionof{figure}{Prompt for Correcting Mistranscription (Part 1).}
\label{fig:prommpt_mt_1}
\end{minipage}

\begin{minipage}{\textwidth}
\begin{tcolorbox}[width=\textwidth]
\footnotesize
\begin{Verbatim}
### Step 3: Rank by Transcript Context

Rank remaining candidates by strength of contextual support in the transcript. Evidence strength, from strongest to weakest:

- An immediately adjacent word that clearly suggests the intended word
- A consistent naming pattern across the full response
- Repeated phonemes or themes in nearby words

If no candidate has meaningful contextual support, return no correction.

### Step 4: Select or Decline

If one candidate is clearly supported by both phonetic similarity and transcript context, propose it as the correction.

If two or more candidates remain equally matched after Steps 1–3, return no correction.

## When Not to Correct

Do not propose a correction when:

- Phonetic evidence is weak or ambiguous.
- No candidate survives the category filter.
- Contextual evidence is weak or diffuse.
- Two or more candidates remain equally plausible after all steps.
- You are uncertain which word was intended.

**Valid English word rule:** If the rejected word is a valid English word, it may reflect the patient's genuine response rather than a transcription error. Only correct at confidence 5.

When in doubt, return no correction.

## Confidence Scale

- 5 = Near-certain transcription error (single phoneme difference, strongly supported by context)
- 4 = Strong phonetic and contextual evidence
- 3 = Plausible but some uncertainty remains
- 2 = Weak evidence — do not correct
- 1 = Highly speculative — do not correct

**Valid English words require confidence 5 to correct. If you cannot reach confidence 5, return no correction.**
\end{Verbatim}
\end{tcolorbox}
\captionof{figure}{Prompt for Correcting Mistranscription (Part 2).}
\label{fig:prommpt_mt_2}
\end{minipage}

\begin{minipage}{\textwidth}
\begin{tcolorbox}[width=\textwidth]
\footnotesize
\begin{Verbatim}

## Worked Example

**Input:**
```json
{
  "R2": {
    "prompt": "Tell me all the words you can that begin with the letter L.",
    "full_response": "love, lug, log, lamp, lunch, lost, lack, back",
    "word_list": ["love", "lug", "log", "lamp", "lunch", "lost", "lack"],
    "rejected_words": ["back"]
  }
}
```

**Reasoning:**
- Step 1: Phonetic candidates for "back" — "lack", "black", "rack", "pack"
- Step 2: Category filter for R2 (must begin with L) — only "lack" survives
- Step 3: "lack" is consistent with surrounding L-words in the response
- Step 4: One candidate remains — propose "lack"

**Output:**
```json
{
  "corrections": {
    "R2": [
      {
        "incorrect_word": "back",
        "correct_word": "lack",
        "confidence": 4,
        "justification": "Only L-candidate; consistent with response pattern"
      }
    ]
  }
}
```
## Output Format
Return only raw JSON. No markdown, no explanation, no additional text.

```json
{
  "corrections": {
    "R1": [
      {
        "incorrect_word": "word",
        "correct_word": "replacement",
        "confidence": 5,
        "justification": "max 7 words"
      }
    ],
    "R2": [],
    "R3": [],
    "R4": []
  }
}
```
Return an empty array for any response with no valid corrections.
\end{Verbatim}
\end{tcolorbox}
\captionof{figure}{Prompt for Correcting Mistranscription (Part 3).}
\label{fig:prommpt_mt_3}
\end{minipage}

\begin{minipage}{\textwidth}
\begin{tcolorbox}[width=\textwidth]
\footnotesize
\begin{Verbatim}
You are an expert in cognitive science and psycholinguistics. You will be given a set of variables derived from verbal fluency tests administered to patients, some of whom may have cognitive impairments. Your task is to propose a causal graph over these variables.

Variables:

CI – Cognitive Impairment: presence of cognitive impairment (e.g., MCI, Alzheimer's disease)
TW – Total Words: total valid words produced in response
WF – Word Frequency: mean corpus frequency of valid words
WL – Word Length: mean character length of valid words
AoA – Age of Acquisition: mean age at which words are typically learned
NC – Number of Clusters: total semantically related word groups
CS – Average Cluster Size: mean words per semantic cluster
SW – Number of Switches: total transitions between semantic clusters
SR – Speech Rate: mean words produced per second

Output only a JSON array of directed edges. Each edge is a two-element array [X, Y] meaning "X is a cause of Y". Use the variable abbreviations above. Do not include any explanation or additional text.

Example format:[["CI", "SR"], ["AoA", "WL"]]
\end{Verbatim}
\end{tcolorbox}
\captionof{figure}{Prompt for Eliciting the Influence Network from the LLM}
\label{fig:prommpt_bn}
\end{minipage}

\newpage\noindent
\section{Feature Discretization Thresholds and Frequencies}
\begin{table}[h]
\centering
    \input{Tables/feature_errors_and_thresholds}
\end{table}
\section{Transcription Error Analysis}
While NeSyQuaKE correctly identifies patterns in the data, the transcriptions may be noisy since pathological speech can be challenging for automatic transcription models.
To estimate the loss due to poor transcriptions, we calculated the percent difference in words present in the manual annotations and the foundation models:
\[\text{Percent Difference} = \frac{|\text{Corrected Count - Original Count}|}{\text{Original Count}}\] 
As shown in Table~\ref{tab:error_analysis}, there is a higher percent difference and significantly greater variability in MCI patients compared to Healthy Controls, particularly in phonemic tasks. These statistics demonstrate the inherent difficulty S2T models face when handling pathological speech, confirming that the primary bottleneck in our pipeline is the acoustic front-end rather than the symbolic grounding logic.

\begin{table}[h]
    \centering
    \input{Tables/tab_error_analysis}
\end{table}

\section{Structure Learning using Peter-Clarke (PC) Algorithm}
We compared the LLM-generated Bayesian Network structure to one obtained by the data-driven Peter-Clarke Algorithm. For this baseline, we learned graphs over the semantic and phonetic features extracted by our pipeline.
In the phonemic task, PC was only able to recover the edge $\text{CI} \rightarrow \text{AoA}$ from the expert graph, and similarly in the semantic task, only the edge $\text{CI} \rightarrow \text{TW}$ was recovered.
\begin{figure}[!h]
    \centering
    %\hfill 
    \begin{minipage}[t]{0.4\textwidth}
        \centering
        \resizebox{0.8\linewidth}{!}{\input{Figures/structure_phonetic_bn}}
        % \captionof{figure}{Expert Graph}
        % \label{fig:expert}
    \end{minipage}
    % \hfill
    \begin{minipage}[t]{0.4\textwidth}
        \centering
        \resizebox{0.8\linewidth}{!}{\input{Figures/structure_semantic_bn}}
    \end{minipage}
    \hfill
    \caption{The \textbf{Influence Networks} over Phonemic (Left) and Semantic features  (Right) learned using PC algorithm representing direct influence relationships between \colorbox{purple!20}{CI}, \colorbox{orange!20}{Lexical}, \colorbox{cyan!20}{Semantic}, and  \colorbox{green!20}{Acoustic} variables.}
    \label{fig:bn_comparison_structure}
\end{figure}
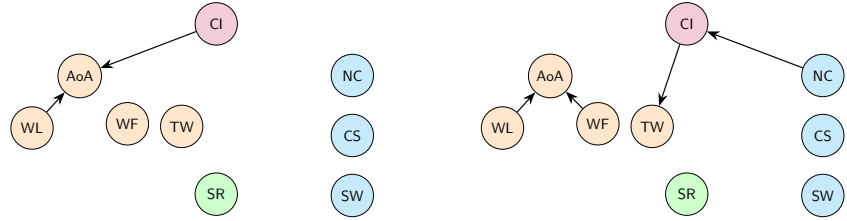
\end{document}

%% file: Tables/features.tex
\caption{\textbf{Variables of Interest.} Lexical, Semantic, and Acoustic variables of interest, along with their descriptions, colored by type: \colorbox{orange!20}{Lexical}, \colorbox{cyan!20}{Semantic}, \colorbox{green!20}{Acoustic}.}
\small
\setlength{\aboverulesep}{0pt}
\setlength{\belowrulesep}{0pt}
\renewcommand{\arraystretch}{1.05}
\begin{tabular}{l l p{7.5cm}}
        \toprule
        \textbf{Type} & \textbf{Variables} & \textbf{Description} \\
        \midrule
        \rowcolor{orange!10} \textbf{Lexical} 
            & Total Words (TW)  & Total valid words in response \\
        \rowcolor{orange!10} & Word Frequency (WF)      & Mean corpus frequency of valid words \\
        \rowcolor{orange!10} & Word Length (WL)         & Mean character length of valid words \\
        \rowcolor{orange!10} & Age of Acquisition (AoA)  & Mean age of learning of valid words\\
        \midrule
        \rowcolor{cyan!10} \textbf{Semantic} 
            & Number of Clusters (NC)   & Total semantically related word groups \\
        \rowcolor{cyan!10} & Average Cluster Size (CS) & Mean number of words per semantic cluster \\
        \rowcolor{cyan!10} & Number of Switches (SW)   & Total transitions between semantic clusters \\
        \midrule
        \rowcolor{green!10} \textbf{Acoustic} & Speech Rate (SR)          & Mean words produced per second \\
        \bottomrule

    \end{tabular}
     \label{tab:features_table_colored}
% \end{table}

% I was technically computing total pause length initially. But would need to change

%% file: Figures/variable_bn.tex
\begin{tikzpicture}[
    node distance=1.2cm and 3cm,
    every node/.style={
        circle, 
        draw=black, 
        minimum size=1cm, 
        font=\small\sffamily,
        align=center,
        fill=white
    },
    >={Stealth[length=3mm]},
    edge/.style={->, thick},
    % Define Pastel Styles
    root/.style={fill=purple!20},      % Cognitive Impairment
    lexical/.style={fill=orange!20},   % TW, WF, WL, AoA
    semantic/.style={fill=cyan!20},    % CS, SW
    acoustic/.style={fill=green!20}    % PL, SR
    ]

    % --- NODES ---
    % Might make better sense for the graph to be horizontal
    % Row 1: Root (Cognitive Type)
    \node[root] (Cog) at (0,0) {CI};
    % Row 2: Level 2 variables
    \node[lexical] (AoA) [below left=0.5cm and 2.5cm of Cog] {AoA};
    \node[lexical] (TW) [below left=1.7cm and 0.1cm of Cog] {TW};
    \node[semantic] (NC) [below right=0.5cm and 2.5cm of Cog] {NC};
    % Row 3: Level 3 variables
    \node[lexical] (WL) [below left=0.5cm and 0.4cm of AoA] {WL};
    \node[lexical] (WF) [below right=0.4cm and 0.4cm of AoA] {WF};
    \node[semantic] (CS) [below=0.4cm of NC] {CS};
    % Row 4: Level 4 variables
    \node[acoustic] (SR) [below=3cm of Cog] {SR};
    \node[semantic] (SW) [below=0.4cm of CS] {SW};
    
    %% Edges Main edges from Cog
    \draw[edge] (Cog) -- (NC);
    \draw[edge] (Cog) to (SW);
    \draw[edge] (Cog) -- (CS);
    \draw[edge] (Cog) -- (TW);
    \draw[edge] (Cog) [bend right = 45] to (WL);
    \draw[edge] (Cog) -- (SR);
    \draw[edge] (Cog) -- (AoA);
    \draw[edge] (Cog) to (WF);
    \draw[edge] (WL) -- (SR);
    \draw[edge] (TW) to (SR);
    \draw[edge] (NC) -- (CS);
    \draw[edge] (CS) -- (SW);
    \draw[edge] (AoA) -- (WF);
    \draw[edge] (AoA) -- (WF);
    \draw[edge] (AoA) -- (WL);
    \draw[edge] (CS) -- (SR);
    
\end{tikzpicture}

%% file: Figures/llm_bn.tex
\begin{tikzpicture}[
    node distance=1.2cm and 3cm,
    every node/.style={
        circle, 
        draw=black, 
        minimum size=1cm, 
        font=\small\sffamily,
        align=center,
        fill=white
    },
    >={Stealth[length=3mm]},
    edge/.style={->, thick},
    % Define Pastel Styles
    root/.style={fill=purple!20},      % Cognitive Impairment
    lexical/.style={fill=orange!20},   % TW, WF, WL, AoA
    semantic/.style={fill=cyan!20},    % CS, SW
    acoustic/.style={fill=green!20}    % PL, SR
    ]

    % --- NODES ---
    % Might make better sense for the graph to be horizontal
    % Row 1: Root (Cognitive Type)
    \node[root] (Cog) at (0,0) {CI};
    % Row 2: Level 2 variables
    \node[lexical] (AoA) [below left=0.5cm and 2.5cm of Cog] {AoA};
    \node[lexical] (TW) [below left=1.7cm and 0.1cm of Cog] {TW};
    \node[semantic] (NC) [below right=0.5cm and 2.5cm of Cog] {NC};
    % Row 3: Level 3 variables
    \node[lexical] (WL) [below left=0.5cm and 0.4cm of AoA] {WL};
    \node[lexical] (WF) [below right=0.4cm and 0.4cm of AoA] {WF};
    \node[semantic] (CS) [below=0.4cm of NC] {CS};
    % Row 4: Level 4 variables
    \node[acoustic] (SR) [below=3cm of Cog] {SR};
    \node[semantic] (SW) [below=0.4cm of CS] {SW};
    
    %% Edges Main edges from Cog
    \draw[edge] (SR) -- (TW);
    \draw[edge] (AoA) to (WL);
    \draw[edge] (AoA) -- (WF);
    \draw[edge] (Cog) -- (AoA);
    \draw[edge] (Cog) to (CS);
    \draw[edge] (Cog) -- (NC);
    \draw[edge] (Cog) -- (SR);
    \draw[edge] (Cog) to (SW);
    \draw[edge] (Cog) -- (TW);
    \draw[edge] (Cog) to (WF);
    \draw[edge] (Cog) -- (WL);
    \draw[edge] (NC) [bend left = 45] to (SW);
    \draw[edge] (NC) -- (TW);
    \draw[edge] (WF) -- (SR);
    \draw[edge] (WF) -- (WL);
    \draw[edge] (WF) -- (TW);
    \draw[edge] (WL) -- (SR);
    \draw[edge] (WL) to [out=270, in=270] (TW);
    \draw[edge] (CS) -- (TW);
    \draw[edge] (CS) -- (SW);
    \draw[edge] (SW) -- (TW);
    
\end{tikzpicture}

%% file: Figures/mi_bar_llm.tex
\begin{minipage}{0.4\textwidth}
    \centering
        \resizebox{\textwidth}{!}{\begin{tikzpicture}
        \begin{axis}[
            xbar=2cm,
            error bars/.cd,
            width=16cm,
            height=9cm,
            enlarge y limits=0.08,
            xlabel={Monotonic Influence Score},
            xlabel style={font=\Large},
            symbolic y coords={WL,WF,SR,NC,TW,AoA,CS,SW},
            ytick=data,
            yticklabel={\detokenize\expandafter{\tick}},
            yticklabel style={font=\Large\bfseries\ttfamily},
            xmin=-0.3, xmax=0.20,
            xtick={-0.3, -0.2,-0.1, 0, 0.1, 0.2},
            xticklabel style={
                font=\Large,
                /pgf/number format/fixed,
                /pgf/number format/precision=2
            },
            xmajorgrids=true,
            grid style={dashed, gray!30},
            xtick pos=left,
            ytick pos=left,
            axis x line*=bottom,
            axis y line*=left,
            bar width=18pt,
        ]
            \addplot[fill=cyan!70!blue, draw=black!70,
    error bars/.cd,
        x dir=both,
        x explicit,] coordinates {
                (-0.24,TW) +- (0.00,0)
                (-0.12,AoA) +- (0.01,0)
                (-0.12,WL) +- (0.01,0)
                (0.16,WF) +- (0.01,0)
                (-0.18,NC) +- (0.01,0)
                (-0.13,CS) +- (0.01,0)
                (-0.18,SW) +- (0.0,0)
                (-0.09,SR) +- (0.01,0)
            };
    \end{axis}
\end{tikzpicture}}
\end{minipage}
\hfill
    \begin{minipage}{0.4\textwidth}
    \centering
        \resizebox{\textwidth}{!}{\begin{tikzpicture}
            \begin{axis}[
                xbar=2cm,
                error bars/.cd,
                width=16cm,
                height=9cm,
                enlarge y limits=0.08,
                xlabel={Monotonic Influence Score},
                xlabel style={font=\Large},
                symbolic y coords={WL,WF,SR,NC,TW,AoA,CS,SW},
                ytick=data,
                yticklabel={\detokenize\expandafter{\tick}},
                yticklabel style={font=\Large\bfseries\ttfamily},
                xmin=-0.20, xmax=0.05,
                xtick={-0.20, -0.15,-0.1, -0.05, 0, 0.05, 0.1, 0.15, 0.2},
                xticklabel style={
                    font=\Large,
                    /pgf/number format/fixed,
                    /pgf/number format/precision=2
                },
                xmajorgrids=true,
                grid style={dashed, gray!30},
                xtick pos=left,
                ytick pos=left,
                axis x line*=bottom,
                axis y line*=left,
                bar width=18pt,
            ]
            \addplot[fill=cyan!70!blue, draw=black!70,
            error bars/.cd,
                x dir=both,
                x explicit,] coordinates {
                (-0.14,TW) +- (0.01,0)
                (-0.12,AoA) +- (0.01,0)
                (0.02,WL) +- (0.01,0)
                (0.02,WF) +- (0.01,0)
                (-0.09,NC) +- (0.01,0)
                (-0.09,CS) +- (0.01,0)
                (-0.13,SW) +- (0.01,0)
                (-0.07,SR) +- (0.01,0)
            };
            \end{axis}
        \end{tikzpicture}}
    \end{minipage}
    \caption{\textbf{Degree of monotonic influence} of cognitive impairment on semantic (left) and phonemic (right) features, as inferred by NeSyQuaKE, averaged over $5$ trials}
    \label{fig:MI_degrees}

%% file: Tables/PCEE.tex
\caption{PCEE values with rule-based extraction vs NeSyQuaKE for \colorbox{orange!20}{Lexical}, \colorbox{cyan!20}{Semantic}, and  \colorbox{green!20}{Acoustic} variables in Semantic and Phonemic Fluency tasks. For NeSyQuaKE and NeSyQuaKE with feedback, the reported values correspond to the mean over five runs. Standard deviations are omitted, as all values were below $10^{-4}$}
\label{tab:pcee_comparison}
\renewcommand{\arraystretch}{1.2}
\resizebox{\textwidth}{!}{%
\begin{tabular}{l ccc ccc}
\hline
\textbf{Variable Name} 
& \multicolumn{3}{c}{\textbf{Semantic Fluency}} 
& \multicolumn{3}{c}{\textbf{Phonemic Fluency}} \\
\cline{2-4} \cline{5-7}
& Rule-based & NeSyQuaKE  & w/Feedback & Rule-based & NeSyQuaKE  & w/Feedback \\ 
\hline
\rowcolor{orange!10} Total Words (TW)          & 0.087 & 0.005 & 0.004 & 0.120 & 0.044 & 0.041\\
\rowcolor{orange!10} Age of Acquisition (AoA)  & 0.003 & 0.001 & 0.001 & 0.003 & 0.001 & 0.001\\
\rowcolor{orange!10} Word Length (WL)          & 0.117 & 0.008 & 0.008 & 0.0 & 0.001 & 0.001\\
\rowcolor{orange!10} Word Frequency (WF)       & 0.005 & 0.002 & 0.002 & 0.004 & 0.001 & 0.001\\ 
\hline

\rowcolor{cyan!10} Number of Clusters (NC)      & 0.117 & 0.006 & 0.004  & 0.057 & 0.012 & 0.009\\
\rowcolor{cyan!10} Average Cluster Size (CS)    & 0.028 & 0.003 & 0.003 & 0.064 & 0.028 & 0.024\\
\rowcolor{cyan!10} Number of Switches (SW)      & 0.059 & 0.002 & 0.001 & 0.133 & 0.030 & 0.022\\ 
\hline

\rowcolor{green!10} Speech Rate (SR)            & 0.070 & 0.002 & 0.002 & 0.162 & 0.007 & 0.007\\ 
\hline

\end{tabular}
}

%% file: Tables/monotonic_influences.tex
\caption{\textbf{NeSyQuaKE recovers expert knowledge.} Comparison of monotonic influence rules extracted by NeSyQuaKE with those elicited from domain experts. Each rule characterizes the influence of Cognitive Impairment (CI) on variables of interest.
% : Total Words (TW), Age of Acquisition (AoA), Word Length (WL), Word Frequency (WF), Number of Clusters (NC), Cluster Size (CS), Number of Switches (SW), and Speech Rate (SR). 
For each source of rules, a ($\checkmark$) indicates the monotonic influence was found,  (\textbf{?}) indicates it was not found, and ($\times$) indicates its opposite was found. An influence statement $\text{A }^\text{M$\pm$}_{\prec}\text{B}$ is read as A has a positive ($M+$) or negative ($M-$) monotonic influence on B.}
\small
\setlength{\aboverulesep}{0pt}
\setlength{\belowrulesep}{0pt}
\renewcommand{\arraystretch}{1.05}
\begin{tabular}{l cc cc}
\hline
\textbf{Rule} 
& \multicolumn{2}{c}{\textbf{Semantic Fluency}} 
& \multicolumn{2}{c}{\textbf{Phonemic Fluency}} \\

\cline{2-3} \cline{4-5}
& Expert & NeSyQuaKE & Expert & NeSyQuaKE \\ 
\hline
\rowcolor{orange!10} $\text{CI }^\text{M-}_{\prec}\text{ TW}$        & ${\checkmark}$ &$\checkmark$& $\textbf{?}$&$\checkmark$\\
\rowcolor{orange!10} $\text{CI }^\text{M-}_{\prec}\text{ AoA}$ & $\checkmark$ &$\checkmark$& $\checkmark$&$\checkmark$\\
\rowcolor{orange!10} $\text{CI }^\text{M-}_{\prec}\text{ WL}$ & $\textbf{?}$ &$\checkmark$  & $\textbf{?}$& $\times$\\
\rowcolor{orange!10} $\text{CI }^\text{M+}_{\prec}\text{WL}$ & $\textbf{?}$ &$\times$ & $\textbf{?}$&$\checkmark$\\
\rowcolor{orange!10} $\text{CI }^\text{M+}_{\prec}\text{WF}$  & $\checkmark$ &$\checkmark$&$\checkmark$ & $\checkmark$\\
\hline
\rowcolor{cyan!10} $\text{CI }^\text{M-}_{\prec}\text{ NC}$     & $\checkmark$ &$\checkmark$&$\textbf{?}$ & $\checkmark$\\ 
\rowcolor{cyan!10} $\text{CI }^\text{M-}_{\prec}\text{ CS}$   & $\textbf{?}$&$\checkmark$ &$\textbf{?}$ & $\checkmark$ \\
\rowcolor{cyan!10} $\text{CI }^\text{M-}_{\prec}\text{ SW}$      & $\checkmark$ & $\checkmark$&$\textbf{?}$&$\checkmark$\\
\hline
\rowcolor{green!10} $\text{CI }^\text{M-}_{\prec}\text{ SR}$           & $\textbf{?}$ & $\checkmark$& $\textbf{?}$& $\checkmark$\\
\hline

\end{tabular}

\label{tab:fluency_indicators}

%TW, NC, SW, WF, AoA - Semantic
%WF, AoA

%% file: Tables/feature_errors_and_thresholds.tex
\caption{\textbf{NeSyQuaKE grounds variables of interest in raw audio recordings.} Discretization Thresholds and Frequencies for \colorbox{orange!20}{Lexical}, \colorbox{cyan!20}{Semantic}, and  \colorbox{green!20}{Acoustic} variables in Semantic and phonemic Fluency tasks.}
\label{tab:variables_combined}
\renewcommand{\arraystretch}{1.2}
\resizebox{.95\textwidth}{!}{%
\begin{tabular}{l cc cc}
\hline
\textbf{Variable Name} 
& \multicolumn{2}{c}{\textbf{Semantic Fluency}} 
& \multicolumn{2}{c}{\textbf{Phonemic Fluency}} \\
\cline{2-3} \cline{4-5}
& Threshold & Frequency& Threshold & Frequency \\ 
\hline
\rowcolor{orange!10} Total Words (TW)          & 23.39 & 83 & 20.97 & 89 \\
\rowcolor{orange!10} Age of Acquisition (AoA)  & 5.30 & 88  & 6.60 & 82 \\
\rowcolor{orange!10} Word Length (WL)          & 5.92 & 90  & 5.30 & 77  \\
\rowcolor{orange!10} Word Frequency (WF)       & 1.7e-5 & 74  & 1.7e-4 & 45 \\ 
\hline

\rowcolor{cyan!10} Number of Clusters (NC)      & 11.88 & 93  & 8.93 & 109\\
\rowcolor{cyan!10} Average Cluster Size (CS)    & 1.06 & 88   & 1.40 & 73  \\
\rowcolor{cyan!10} Number of Switches (SW)      & 14.39 & 77 & 14.50 & 82 \\ 
\hline

\rowcolor{green!10} Speech Rate (SR)            & 0.35 & 62   & 0.28 & 77  \\ 
\hline

\end{tabular}
}

%% file: Tables/tab_error_analysis.tex
\caption{Semantic and Phonetic Fluency Word Counts Before and After Correction}
%\resizebox{\linewidth}{!}{%
\begin{tabular}{llccc}
\toprule
\textbf{Task} & \textbf{Group} & \textbf{Original} & \textbf{Corrected}  & \textbf{Percent Difference} \\
\midrule
\multirow{2}{*}{Semantic} & 
HC & 35.62 $\pm$ 7.20 & 35.82 $\pm$ 6.96 & $3.94\%\pm 4.64 \% $\\
& MCI & 26.58 $\pm$ 8.54 & 27.44 $\pm$ 8.41 & $9.08\% \pm 10.73\%$\\
\midrule
\multirow{2}{*}{Phonemic}
 & HC & 28.30 $\pm$ 7.94 & 29.69 $\pm$ 7.46 & $8.41\% \pm 15.20\%$ \\
 & MCI & 22.24 $\pm$ 9.26 & 24.37 $\pm$ 8.69 & $18.04\%\pm 20.74\%$ \\
\bottomrule
\label{tab:error_analysis}
\end{tabular}
%}

%% file: Figures/structure_phonetic_bn.tex
\begin{tikzpicture}[
    node distance=1.2cm and 3cm,
    every node/.style={
        circle,
        draw=black,
        minimum size=1cm,
        font=\small\sffamily,
        align=center,
        fill=white
    },
    >={Stealth[length=3mm]},
    edge/.style={->, thick},
    edge_label/.style={
        midway,
        fill=white,
        inner sep=1pt,
        font=\scriptsize\sffamily
    },
    root/.style={fill=purple!20},
    lexical/.style={fill=orange!20},
    semantic/.style={fill=cyan!20},
    acoustic/.style={fill=green!20}
    ]

    % --- NODES ---
    \node[root] (CI) at (0,0) {CI};

    \node[lexical] (AoA) [below left=0.5cm and 2.5cm of CI] {AoA};
    \node[lexical] (TW) [below left=1.7cm and 0.1cm of CI] {TW};
    \node[semantic] (NC) [below right=0.5cm and 2.5cm of CI] {NC};

    \node[lexical] (WL) [below left=0.5cm and 0.4cm of AoA] {WL};
    \node[lexical] (WF) [below right=0.4cm and 0.4cm of AoA] {WF};
    \node[semantic] (CS) [below=0.4cm of NC] {CS};

    \node[acoustic] (SR) [below=3cm of CI] {SR};
    \node[semantic] (SW) [below=0.4cm of CS] {SW};

    % --- POOLED AVERAGE PHONEMIC DAG EDGES ---
    \draw[edge] (CI) -- (AoA);
    \draw[edge] (WL) -- (AoA);

\end{tikzpicture}

%% file: Figures/structure_semantic_bn.tex
\begin{tikzpicture}[
    node distance=1.2cm and 3cm,
    every node/.style={
        circle,
        draw=black,
        minimum size=1cm,
        font=\small\sffamily,
        align=center,
        fill=white
    },
    >={Stealth[length=3mm]},
    edge/.style={->, thick},
    edge_label/.style={
        midway,
        fill=white,
        inner sep=1pt,
        font=\scriptsize\sffamily
    },
    root/.style={fill=purple!20},
    lexical/.style={fill=orange!20},
    semantic/.style={fill=cyan!20},
    acoustic/.style={fill=green!20}
    ]

    % --- NODES ---
    \node[root] (CI) at (0,0) {CI};

    \node[lexical] (AoA) [below left=0.5cm and 2.5cm of CI] {AoA};
    \node[lexical] (TW) [below left=1.7cm and 0.1cm of CI] {TW};
    \node[semantic] (NC) [below right=0.5cm and 2.5cm of CI] {NC};

    \node[lexical] (WL) [below left=0.5cm and 0.4cm of AoA] {WL};
    \node[lexical] (WF) [below right=0.4cm and 0.4cm of AoA] {WF};
    \node[semantic] (CS) [below=0.4cm of NC] {CS};

    \node[acoustic] (SR) [below=3cm of CI] {SR};
    \node[semantic] (SW) [below=0.4cm of CS] {SW};

    % --- POOLED AVERAGE SEMANTIC DAG EDGES ---
    \draw[edge] (CI) -- (TW);
    \draw[edge] (NC) -- (CI);
    \draw[edge] (WF) -- (AoA);
    \draw[edge] (WL) -- (AoA);

\end{tikzpicture}